\documentclass{article}

 \usepackage[preprint]{neurips_2025}

\usepackage[utf8]{inputenc} 
\usepackage[T1]{fontenc}    
\usepackage{hyperref}       
\usepackage{url}            
\usepackage{booktabs}       
\usepackage{amsfonts}       
\usepackage{nicefrac}       
\usepackage{microtype}      
\usepackage{xcolor}         

\usepackage{microtype}
\usepackage{graphicx}
\usepackage{subfigure}

\usepackage{amsmath}
\usepackage{amssymb}
\usepackage{mathtools}
\usepackage{amsthm}

\usepackage[capitalize,noabbrev]{cleveref}
\title{Evaluating the Safety and Skill Reasoning of Large Reasoning Models Under Compute Constraints}

\author{%
 Adarsha Balaji\\
 Mathematics and Computer Science Division\\
 Argonne National Laboratory\\
 Lemont, IL 60439 \\
  \texttt{abalaji@anl.gov} \\
   \And
   Le Chen \\
   Data Science and Learning Division\\
   Argonne National Laboratory\\
   Lemont, IL 60439 \\
   \AND  
   Rajeev Thakur \\
   Data Science and Learning Division\\
   Argonne National Laboratory\\
   Lemont, IL 60439 \\
   \And   
   Franck Cappello \\
    Mathematics and Computer Science Division \\
  Argonne National Laboratory\\
  Lemont, IL 60439 \\ 
 \And  
  Sandeep Madireddy \\
  Mathematics and Computer Science Division \\
  Argonne National Laboratory\\
  Lemont, IL 60439 \\
  \texttt{smadireddy@anl.gov} \\
}

\begin{document}

\maketitle

\begin{abstract}
Test-time compute scaling has demonstrated the ability to improve the performance of reasoning language models by generating longer chain-of-thought (CoT) sequences. However, this increase in performance comes with a significant increase in computational cost. In this work, we investigate two compute constraint strategies: (1) reasoning length constraint and (2) model quantization, as methods to reduce the compute demand of reasoning models and study their impact on their safety performance. Specifically, we explore two approaches to apply compute constraints to reasoning models: (1) fine-tuning reasoning models using a length-controlled policy optimization (LCPO) based reinforcement learning method to satisfy a user-defined CoT reasoning length, and (2) applying quantization to maximize the generation of CoT sequences within a user-defined compute constraint. Furthermore, we study the trade-off between the computational efficiency and the safety of the model. 
\end{abstract}

\section{Introduction}

Existing benchmarks and evaluation protocols fall short of capturing the trade-offs between compute at test-time i.e. reasoning length and performance i.e. accuracy and safety of large reasoning models (LRMs). First, they often conflate raw performance with computational cost \cite{xu2025towards}, either by increasing model size or by increasing their reasoning budget at inference time \cite{snell2024scaling}. For example, a leaderboard for math problem solving might rank models solely by final answer accuracy, implicitly rewarding those that secretly use extremely long chain-of-thought or multiple sample voting to get a higher score. Such improvements typically come with higher computational costs, often scaling with the size (parameters) of LRMs.


In practice, such performance differences matter at inference-time, yet evaluations rarely report the average token usage or compute time per question. Only recently have reasoning models like S1 \cite{muennighoff2025s1} and L1 \cite{aggarwal2025l1} begun to emphasize efficiency metrics such as accuracy as a function of compute budget (tokens). Without these, researchers risk pursuing methods that yield marginal accuracy gains at disproportionate compute costs. Moreover, performance is often tied to the number of steps a model takes to complete a task without considering its size (parameters) or the number of operations per step (FLOPs). For instance, GPT-4, a  1.76 trillion parameters model, might solve a puzzle in 50 steps whereas a much smaller model might need 500; the compute budgets for the two models completing the same task are vastly different. In addition to model size, quantization can often balance the efficiency and performance by reducing the compute and memory footprint of a model with minimal loss in performance. Therefore, there is a need to evaluate the performance of reasoning models as a function of both accuracy and compute efficiency, rather than accuracy alone.


While compute constraints such as limited reasoning budgets and quantization have been studied for general, science, and math reasoning, their impact on model safety remains unexplored.

In summary, our contributions are: 

\begin{itemize}
\item We study the impact of test-time compute-constraint methods on both skill and safety of LRMs;
\item We use the Length Controlled Policy Optimization reinforcement learning method, presented in \cite{aggarwal2025l1}, to safety fine-tune a reasoning model with a precise user-defined length control using the SafeChain dataset \cite{jiang2025safechain};
    \item We apply weight quantization (GPTQ) methods on the baseline and safety fine-tuned models and study its impact on skill and safety performance for increasing reasoning chain-of-thought lengths;
\item We analyze the trade-off between the two compute-constraint methods under a fixed compute budget.
\end{itemize}

\section{Related Works}
Test time compute performance scales with increased compute - the more reasoning tokens a model can generate, the more accurate its response \cite{rosenfeld2019constructive}. 

\begin{figure}[t!]
	\centering
	\centerline{\includegraphics[width=2.2in, height=1.5in]{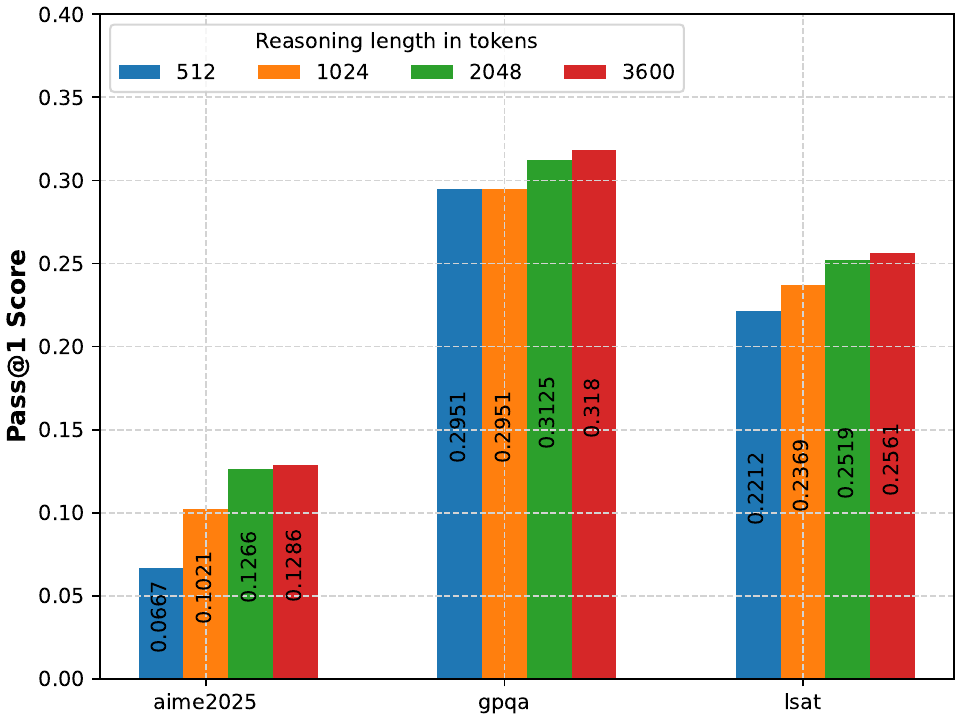}}
	\caption{Performance of the baseline L1 model for science and math reasoning skills for increasing reasoning length.}
	\label{fig:fp-skill-l1}
\end{figure}

\paragraph{Reasoning Models.} The emergence of Large Reasoning Models (LRMs) reflects a shift from treating reasoning as an incidental property of LLMs to deliberately training models to “think before they answer.” Early methods such as Chain-of-Thought (CoT) prompting~\cite{wei2022chain}, self-consistency, and more structured approaches like Tree- and Graph-of-Thought~\cite{yao2023tree, besta2024graph} demonstrated that extending intermediate reasoning can improve accuracy but at increasing computational cost~\cite{chen2024landscape}. Building on this, LRMs such as OpenAI’s o1~\cite{jaech2024openai} and DeepSeek-R1~\cite{guo2025deepseek} employ large-scale supervised fine-tuning and reinforcement learning to explicitly incentivize the generation of long, structured reasoning traces. This paradigm enables models to decompose complex tasks, explore alternatives, and self-correct, but also tightly couples performance to test-time computation. The reliance of LRMs on extended reasoning motivates our study of how compute constraints through length control or quantization impact their skill and safety trade-offs in practice.

\paragraph{Quantization}
Quantization is an approach that reduces the precision of model parameters like weights and activations. It is common for neural networks to be implemented in bfloat16, a format supported by most neural accelerators.  
Post-training quantization (PTQ) approaches such as GPTQ \cite{frantar2022gptq}, GGUF \cite{lin2016neural}, and ZeroQuant \cite{yao2022zeroquant} enable efficient compression without retraining, while quantization-aware training (QAT) methods such as LR-QAT \cite{bondarenko2024low} preserve task-specific performance through learnable scaling factors. Recent work extends these techniques to LLMs, though primarily evaluated on language understanding rather than reasoning tasks.

\section{Method}

\subsection{Safety Reasoning Length Control}
In this work, we fine-tune the baseline reasoning model to improve its safety performance while conditioning the model on a user-defined reasoning token length. We extend the work done in \cite{aggarwal2025l1} and use the length controlled policy optimization (LCPO) reinforcement learning method presented in \cite{aggarwal2025l1} to fine-tune our model. We modify the reward function to combine the (1) safety reward and (2) the length penalty. The safety reward is determined by using the Llama-Guard-3 model.

\subsubsection{Dataset Creation}
We conduct our training on the L1-Exact-1.5B model \cite{aggarwal2025l1} using a chain-of-thought (CoT) style safety dataset called SafeChain \cite{jiang2025safechain} that can improve a model's safety performance while preserving its math and coding performance across all benchmarks.

To enable length control, we augment each prompt in the SafeChain dataset with a target length instruction.

\begin{equation}
    X_i = x_i + \text{"Think for n tokens"}
\end{equation}

where, $x_i$ is the prompt from the SafeChain dataset and $X_i$ is the augmented prompt used for training.
The value n is randomly picked from 0 to 4000 for each prompt.

\subsection{Quantization}
We conduct a comprehensive investigation into the effects of quantization techniques, in particular weight-only quantization (GPTQ). GPTQ \cite{frantar2022gptq} is a one-shot weight quantization method,
based on approximate second-order information and error compensation, that is both highly-accurate and highly-efficient. We perform INT8 (8-bit) and INT4 (4-bit) quantization of the weights of the model while leaving the activation operation at the original 16-bit.

\section{Evaluation}\label{sec:exp}
\subsection{Evaluated Benchmarks}
We evaluate the compute-constrained models using science, math, and safety representative reasoning benchmarks: GPQA Diamond \cite{rein2024gpqa} consists of 198 PhD-level science questions from Biology, Chemistry, and Physics. MATH500 \cite{hendrycks2021measuring} is a benchmark consisting of competition-level math problems of varying difficulty. AIME2025 \cite{hendrycks2020measuring} contains 30 problems from the AIME1 and AIME2 math jam. Following previous work \cite{aggarwal2025l1}, we evaluate our model on the same subset selected by OpenAI \cite{achiam2023gpt}. In addition to these three common science and math reasoning benchmarks, we evaluated the safety performance of the reasoning models using the StrongReject dataset \cite{souly2024strongreject}, a state-of-the-art safety evaluation dataset with 60 jailbreak queries. 
For all benchmarks, we generate a sample for each question with a temperature of 0 (greedy) to measure accuracy. Through these benchmarks, we can evaluate the reasoning ability of LLMs from different perspectives.

\subsection{Safety Evaluator}
Our safety evaluation work builds on the prior work \cite{jaech2024openai, jiang2025safechain}. We consider the LLama-Guard \cite{chi2024llama} evaluator to generate a safety score based on the work done in \cite{jiang2025safechain} to assess the effectiveness of \textit{four} state-of-the-art safety evaluators - Llama-guard \cite{chi2024llama}, Refusal String Matching (RS-Matching) \cite{zou2023universal}, OpenAI Moderation API \cite{openAI-API}, and HarmBench \cite{mazeika2024harmbench}.

\subsection{Metric}

\subsubsection{Skill Evaluation}
\label{sec:supp-skill}
We use the pass@1 metric to evaluate the skill and safety performance of the reasoning models discussed in this work. We define the pass@1 score as shown in Equation~\ref{eq:pass1}: 

\vspace{-0.15cm}
\begin{equation}
	\label{eq:pass1}
	pass@1 = \frac{1}{K} \cdot \sum_{i=1}^{K}{p_{i}} 
\end{equation}
\vspace{-0.15cm}    

where $p_{i}$ is a binary score that indicates whether a response $y_{i}$ to a query $q_i$ is correct for skill tasks.

\subsubsection{Safety Evaluation}
\label{sec:supp-safty}
We use the safe@1 metric to evaluate the skill and safety performance of the reasoning models discussed in this work. As described in \cite{guo2025deepseek, jiang2025safechain} we define the safe@1 score as shown in Equation~\ref{eq:safe1}: 

\vspace{-0.15cm}
\begin{equation}
	\label{eq:safe1}
	safe@1 = \frac{1}{K} \cdot \sum_{i=1}^{K}{s_{i}} 
\end{equation}
\vspace{-0.15cm}    

where $s_{i}$ is a binary score, generated using a state-of-the-art evaluator, that indicates whether a response $y_{i}$ to a query $q_i$ is correct for skill tasks and safe (1) or not (0) for safety tasks. We generate the safe score (s) using the Llama-Guard-3-8B \cite{chi2024llama} evaluator.

\subsection{Evaluation Protocol}
We evaluate our compute constraint approaches using the following methods: (1) we evaluate the overall performance (skill and safety reasoning) when generating responses at different target lengths. In our experiments, target lengths are selected from {512, 1024, 2048, 3600} tokens; (2) we evaluate the overall performance (skill and safety reasoning) of the quantized models when generating responses at different target lengths.

\subsection{Reasoning Length Controlled Compute-Constraint}

\subsubsection{Science and Math Skill Evaluation}
We start by evaluating the baseline L1 model using science and math skill datasets. We choose the AIME, GPQA, and LSAT reasoning datasets to evaluate L1 as they have not been used in the training of the L1 model. Figure \ref{fig:fp-skill-l1} shows the performance of the L1 model for an increasing number of reasoning tokens (512, 1024, 2048, and 3600). We observe that the reasoning performance of the L1 model scales with an increase in reasoning tokens used to generate an answer. While the reasoning performance scales linearly for the GPQA and LSAT datasets, we observe that the evaluation on the AIME dataset scales well for reasoning tokens below 2048 tokens and evens out for larger reasoning lengths. The key trend we want to highlight here is that an {\it increase in reasoning length increases the performance of these models}.

\section{Results}

\subsection{Baseline}
We evaluate our compute-constrained reasoning model against the following baseline models:

\begin{itemize}
\item L1-Qwen-1.5B: a LCPO-based fined-tuned version of Agentic-24K with a context length of 4K. The model serves as a fair baseline for a reasoning length controlled model. For brevity, we refer to this model as L1-1.5B.
\item L1-Qwen-8B: a LCPO-based fined-tuned version of Agentic-24K with a context length of 4K. The model serves as a fair baseline for a reasoning length constrained model. For brevity, we refer to this model as L1-8B.
\end{itemize}

\subsection{Reasoning Length Controlled Compute-Constraint}

\subsubsection{Science and Math Skill Evaluation}
We start by evaluating the baseline L1 model using science and math skill datasets. We choose the AIME, GPQA, and LSAT reasoning datasets to evaluate L1 as they have not been used in the training of the L1 model.  

\begin{figure*}[h!]
	\centering
    \vspace{-10pt}
	\centerline{\includegraphics[width=5.5in, height=1.5in]{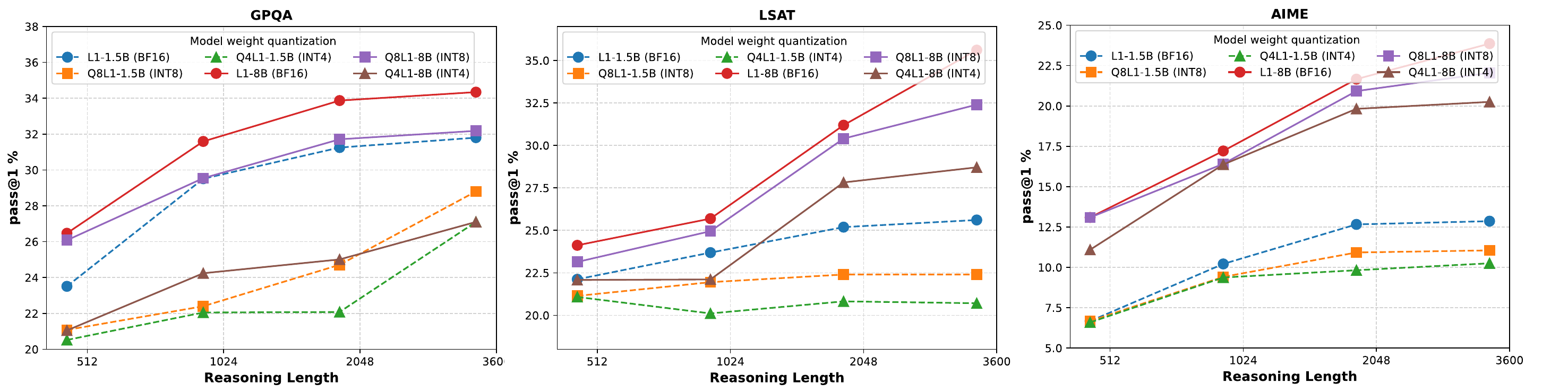}}
	\caption{Performance of the baseline L1 and its post-training quantized models for skill based reasoning tasks - (a) GPQA, (b) LSAT and (c) AIME.}
    \vspace{-10pt}
	\label{fig:quantized-l1}
\end{figure*}

Figure \ref{fig:quantized-l1} shows the performance of the L1 model for an increasing number of reasoning tokens (512, 1024, 2048, and 3600). We observe that the reasoning performance of the L1 model scales with an increase in reasoning tokens used to generate an answer. While the reasoning performance scales linearly for the GPQA and LSAT datasets, we observe that the evaluation on the AIME dataset scales well for reasoning tokens below 2048 tokens and evens out for larger reasoning lengths. The key trend we want to highlight here is that an {\it increase in reasoning length increases the performance of LRMs}.  

\begin{figure}[h!]
	\centering
    \vspace{-10pt}
	\centerline{\includegraphics[width=3.7in, height=1.5in]{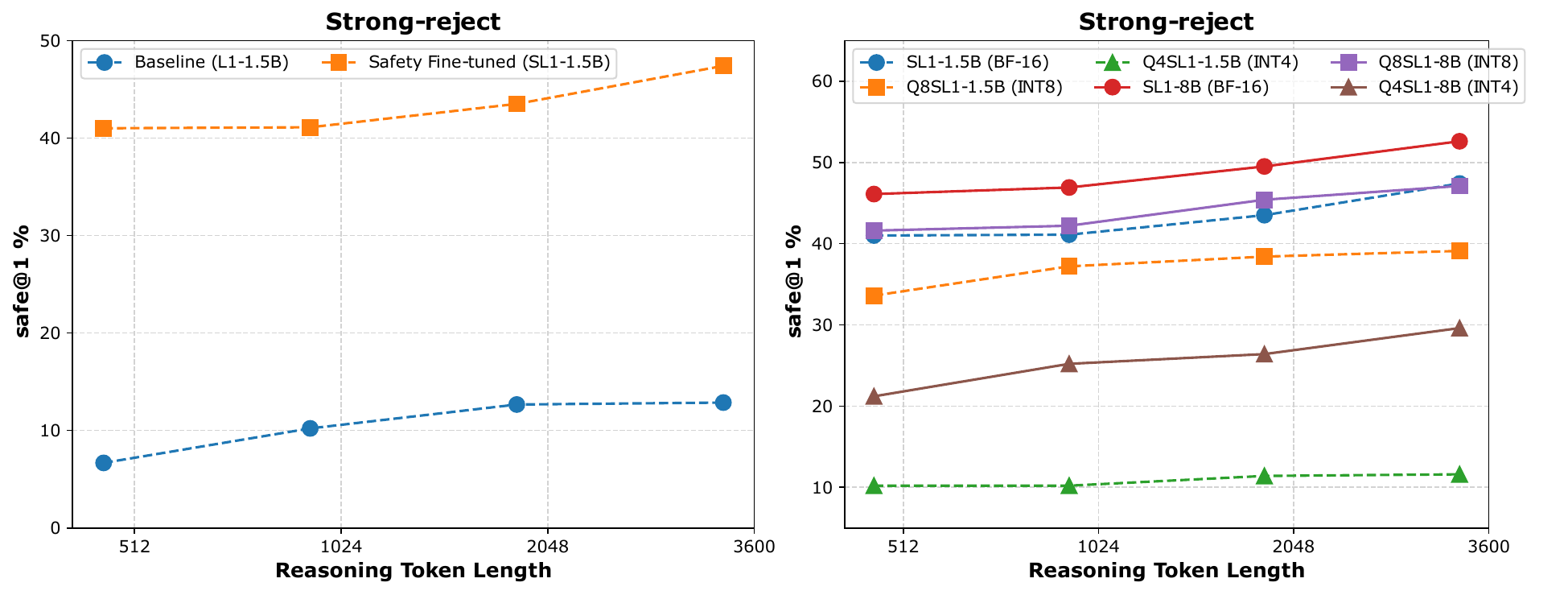}}
	\caption{(a) Performance of the baseline L1 and SL1 (Safety fine-tuned L1) under varying reasoning token budgets; (b) performance of SL1-1.5B and SL1-8B models for BF16 (circle), INT8 (square) and INT4 (triangle) weight precision.}
    \vspace{-10pt}
	\label{fig:fp-safety-l1}
\end{figure}

\subsubsection{Safety Evaluation}
In this section, we quantitatively assess the safety reasoning capability of the baseline L1 model. Figure \ref{fig:fp-safety-l1} (a) illustrates the safety performance of the baseline L1-1.5B model when evaluated using the StrongReject \cite{souly2024strongreject} dataset for increasing reasoning length. The safe@1 score (defined in Section~\ref{sec:supp-safty}) indicates the percentage of safe (1) or unsafe (0) responses to jailbreak queries from the StrongReject dataset. This score is generated using the Llama-Guard-3 \cite{chi2024llama} safety evaluator. 

From Figure \ref{fig:fp-safety-l1} (a) we observe that the safety performance of the baseline L1-1.5B model does not match the state-of-the-art safety performance of similar 1.5B models \cite{jiang2025safechain}. This is expected, as the baseline L1 model has been finetuned using only science and math reasoning skill-based datasets. To address this poor safety performance, we fine-tune the baseline L1-1.5B and L1-8B models (S-L1) using the LCPO RL method and the Safechain \cite{jiang2025safechain} dataset. Training is performed for 300 iterations using the VeRL engine. We train the model using reasoning traces with target lengths between 1 and 4000 tokens. Figure \ref{fig:fp-safety-l1} (a) and (b) (circle) illustrate the improved safety performance of the safety fine-tuned L1-1.5B and L1-8B models. For the rest of this work, we refer to safety fine-tuned models as SL1.

\subsection{Quantization Compute-Constraint}
In this section, we study the impact of quantization-based compute constraints on the performance of reasoning models. We evaluate the reasoning models for skill and safety performance. 

\subsubsection{Science and Math Skill Evaluation}
Figure \ref{fig:quantized-l1} illustrates the performance of the post-training weight quantized L1 model for LSAT, AIME and GPQA reasoning datasets. We compare the performance of the full-precision L1 model with \textit{two} levels of weight quantization - INT8 and INT4. Using post-training quantization method (GPTQ), we create the Q8L1 (INT8) and Q4L1 (INT4) models.

We make \textit{two} observations: (1) the reasoning performance of the L1, Q8L1, and Q4L1 models improves with an increase in reasoning length, but this effect is less profound as we increase the quantization level from INT8 to INT4, and (2) the performance of the INT4 quantized model drops significantly for all three evaluated skill datasets, irrespective of the reasoning length. Due to observation \textit{two}, we limit our study to INT8 quantized models. 

\subsubsection{Safety Evaluation}
Figure \ref{fig:fp-safety-l1} B illustrates the safety performance of the SL1-1.5B and SL1-8B models in comparison to its INT8 and INT4 weight quantized implementations. \textit{First}, observe that the safety performance of the baseline and quantized models improves with an increase in reasoning length. \textit{Second}, we observe a significant drop in performance of the 4-bit quantized model (SL1) for safety. For a reasoning length of 512, the safe@1 score drops from ~40\% to ~10\%, making the quantized model far more susceptible to jailbreak queries. However, the safety performance of the 8-bit quantized model does not deteriorate significantly. We observe a 3-7\% drop in safety performance of the 8-bit quantized models when compared to the baseline.

\begin{figure}[h!]
	\centering
	\vspace{-5pt}
	\centerline{\includegraphics[width=4.8in, height=2.7in]{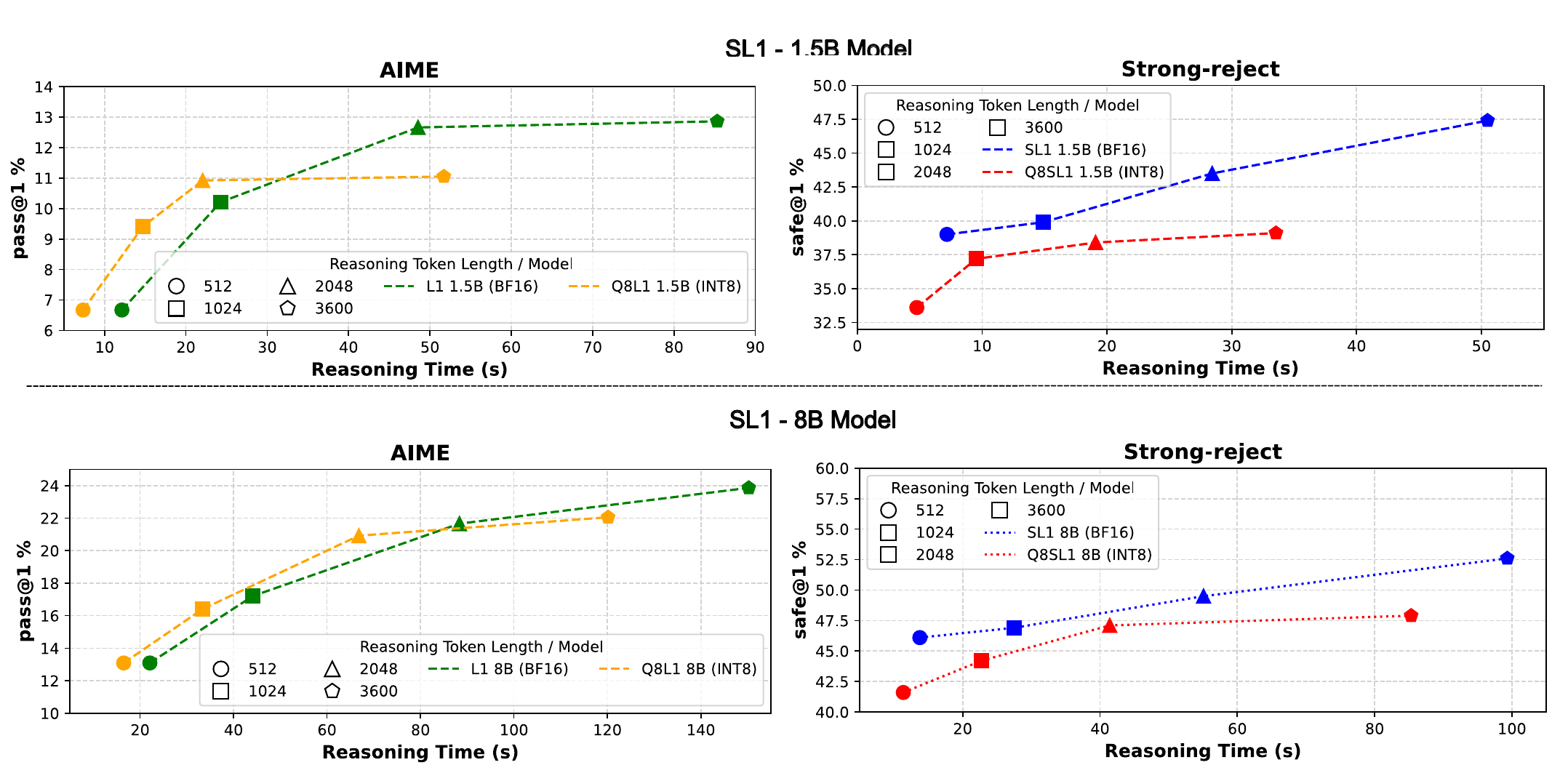}}
 	\caption{Trade-off between compute constraint methods with a fixed compute budget. (a) Evaluation of the SL1 model (BF16 and INT8) using the AIME dataset; (b) Evaluation of the SL1 model (BF16 and INT8) using the StrongReject dataset.}
    \vspace{-10pt}
	\label{fig:compare}
\end{figure}

\begin{table}[t!]
\renewcommand{\arraystretch}{1.2}
\setlength{\tabcolsep}{6pt}
\centering
{\fontsize{8}{10}\selectfont
\begin{tabular}{lcc}
\toprule
Model        & AIME (tokens/s) & StrongReject (tokens/s) \\
\midrule
SL-1.5B      & 42.11  & 71.60  \\
Q8-SL-1.5B   & 69.19  & 107.33 \\
SL-8B        & 23.18  & 31.04  \\
Q8-SL-8B     & 37.21  & 45.10  \\
\bottomrule
\end{tabular}}
\caption{Average throughput (tokens/s) of the evaluated models on the AIME and StrongReject datasets.}
\label{tab:throughput}
\end{table}

\subsection{Impact of Compute Constraints on Reasoning }
In this section, we study the impact of the two chosen compute constraint methods on skill and safety performance. We also detail the trade-off between the two methods, with an aim to demonstrate that full precision and quantized models can show similar performance with-in a fixed compute budget -by varying their reasoning token lengths. To relate the performance of the model to its compute budget, we propose observing the reasoning time, i.e., the inference time (seconds) for the model to generate reasoning tokens. This metric combines the throughput (tokens/s) of a model with the number of reasoning tokens it is afforded within the compute budget.

We report the throughput (tokens/s) of the SL1 and Q8SL1 models in Table \ref{tab:throughput}. The throughput is measured separately (average length of the queries (tokens) varies) for a batch size of 1 on an A100 GPU with a GPU utilization of 0.6. Details on the experimental setup used can be found in Section \ref{sec:exp}. Figure \ref{fig:compare} illustrates the performance of the full precision and quantized versions of SL1 model. The skill and safety reasoning performance of the model is measured as a function of the total compute (tokens) needed to generate an answer. The compute budget is measured as the reasoning time in seconds (x-axis) needed to generate a fixed number of tokens (512, 1024, 2048, and 3600 in this experiment). We calculate the compute budget by multiplying the throughput (tokens/s) of the model for a given dataset by the total number of reasoning tokens.

In Figure \ref{fig:compare} (column one) we observe the evaluation of L1 and Q8L1 models using the AIME dataset. For a reasoning length of 512 tokens, the L1 and QL1 models have a similar accuracy. However, the reasoning time (seconds) of the Q8L1-1.5B model of 7.355 seconds is 39.32\% lower than the reasoning time of the L1 (BF16) model at 12.13 seconds. This is also true for a reasoning length of 1024 tokens. In the case of a reasoning length of 2048 tokens, the L1 model has a higher pass@1 accuracy of 12.66\% when compared to the B8SL1 model with 10.92\%. However, we observe that the SL1 model reasons for 48.54 seconds, a 119.98\% increase in compute budget when compared to the B8L1 model at 22.06 seconds. In figure \ref{fig:compare} (a), we highlight that the L1 model reasoning for 1024 tokens (green square) has a similar performance (1\%) to a Q8SL1 model reasoning for 2048 tokens (orange triangle) with a similar compute budget. Similarly, in figure \ref{fig:compare} (c) we highlight that the L1 model reasoning for 1024 tokens (green square) has a similar performance (1\%) to a Q8SL1 model reasoning for 1024 tokens (orange triangle) with a 16.66\% smaller compute budget.

In Figure \ref{fig:compare} (a) we see the evaluation of SL1-1.5B and Q8SL1-1.5B models using the StrongReject dataset. We observe that the QSL1 model reasoning for 2048 tokens (red triangle) has a safety performance similar (1.4\% drop in safety score) to the SL1 model reasoning for 1024 tokens (blue square). In this scenario, both the models even have a similar compute budget of 19.08 seconds and 14.91 seconds for the Q8SL1 and SL1 models, respectively. This is further highlighted with the SL1-8B and Q8SL1-8B models. We observe that the QSL1-8B model for a reasoning length of 2048 (triangle)  has a safety performance similar (2\% drop in safety score) to the SL1-8B model reasoning for 2048 tokens (blue square) with a compute budget 16.4\% lower.

\vspace{-0.3cm}
\section{Conclusions}

\vspace{-0.3cm}
In this work, we study how compute constraints affect the safety performance of reasoning models. We explore two methods to apply compute constraints: (1) a Length Controlled Policy Optimization (LCPO), a simple reinforcement learning-based method that enables user-defined control over reasoning length, and (2) weight quantization, which reduces the compute demands of the reasoning model and ensures their execution within a user-defined compute budget, namely, inference time. We further demonstrate that within a fixed compute budget (reasoning time), a quantized reasoning model can perform at par with a full-precision model. This is because within the fixed compute budget, the quantized model can generate more reasoning tokens and hence compensate for the loss in performance observed due to quantization.  

\bibliography{neurips_2025}
\bibliographystyle{neurips}

\end{document}